\documentclass[letterpaper, 10 pt, conference]{ieeeconf}
\IEEEoverridecommandlockouts
\usepackage{etoolbox}
\makeatletter
\patchcmd{\@makecaption}
  {\scshape}
  {}
  {}
  {}
\makeatother

\usepackage{color}
\usepackage{graphics} 
\usepackage{epsfig} 
\usepackage{mathptmx} 
\usepackage{times} 
\usepackage{amsmath} 
\usepackage{amssymb}  
\usepackage{graphicx}
\usepackage{multirow, makecell}
\usepackage{array}
\usepackage{booktabs} 
\usepackage{comment}
\usepackage[linesnumbered,ruled]{algorithm2e} 
\newcolumntype{M}[1]{>{\centering\arraybackslash}m{#1}}
\newcommand\T{\rule{0pt}{1.0ex}}       
\newcommand\B{\rule[-1.0ex]{0pt}{0pt}} 

\title{\LARGE \bf
Impact of Traffic Lights on Trajectory Forecasting of Human-driven Vehicles Near Signalized Intersections}

\author{Geunseob (GS) Oh, Huei Peng
\thanks{Geunseob (GS) Oh and Huei Peng are with the Department of Mechanical Engineering, University of Michigan, Ann Arbor, MI 48109 USA, Email : \{gsoh, hpeng\}@umich.edu}
}

\begin{document}

\maketitle
\thispagestyle{empty}
\pagestyle{empty}

\begin{abstract}
Forecasting trajectories of human-driven vehicles is a crucial problem in autonomous driving. Trajectory forecasting in the urban area is particularly hard due to complex interactions with cars and pedestrians, and traffic lights (TLs). Unlike the former that has been widely studied, the impact of TLs on the trajectory prediction has been rarely discussed. In this work, we first identify the less studied, perhaps overlooked impact of TLs. Second, we present a novel resolution that is mindful of the impact, inspired by the fact that human drives differently depending on signal phase (green, yellow, red) and timing (elapsed time). Central to the proposed approach is \emph{Human Policy Models} which model how drivers react to various states of TLs by mapping a sequence of states of vehicles and TLs to a subsequent action (acceleration) of the vehicle. We then combine the Human Policy Models with a known transition function (system dynamics) to conduct a sequential prediction; thus our approach is viewed as \emph{Behavior Cloning}. One novelty of our approach is the use of vehicle-to-infrastructure communications to obtain the future states of TLs. We demonstrate the impact of TL and the proposed approach using an ablation study for longitudinal trajectory forecasting tasks on real-world driving data recorded near a signalized intersection. Finally, we propose \emph{probabilistic} (generative) Human Policy Models which provide probabilistic contexts and capture competing policies, e.g., \textit{pass} or \textit{stop} in the yellow-light dilemma zone.
\end{abstract}


\section{Introduction}
\label{sec:intro}
Autonomous driving has been more successful in highway than in urban city mainly due to the simplicity of its driving environment; absence of traffic signals, and more stable interactions with other vehicles. Realizing fully autonomous vehicles in urban driving environments is more challenging for the opposite reasons.

One of the major differences between urban city and highway driving is traffic lights (TLs). In urban areas, especially in the vicinity of TLs exemplified by signalized corridors or intersections, the motions of vehicles are mainly governed by traffic signals. People obey the traffic signals and properly respond to implicit rules imposed by traffic lights. Examples of the implicit traffic rules include stopping for a traffic light in a red phase, maintaining a proper speed in a green phase in a free-flow situation. This is why predicting how human drivers respond to traffic signals is the key to the trajectory forecasting in urban area. The decision-making, path planning, and control synthesis all benefit from more accurate trajectory forecasting, ultimately leading to successful autonomous driving.

Recent studies in trajectory forecasting utilize generative models (e.g., variational autoencoders (VAE) \cite{ref:2017_Desire} or generative adversarial networks (GAN) \cite{ref:2018_SGAN, ref:2019_Intention}) or convolutional \& recurrent neural-nets models \cite{ref:2018_R1_2, ref:2019_R1_4, ref:2019_R1_5, ref:2016_SocialLSTM, ref:2019_Traphic, ref:2018_R1_1}. The existing works mainly focus on accurate modeling of the interactions among vehicles \cite{ref:2017_Desire, ref:2018_R1_2, ref:2019_R1_4, ref:2019_R1_5, ref:2019_R1_3}, and/or pedestrians \cite{ref:2018_SGAN, ref:2016_SocialLSTM, ref:2019_Traphic, ref:2018_R1_1}. Despite the important role of TLs in the vehicle motions, efforts to understand the dynamics between TLs and vehicle trajectories and to quantify the impacts of TLs on the trajectories have barely been made in the trajectory forecasting literature. 

\begin{figure}[t] 
    \centering
    \includegraphics[width=\linewidth]{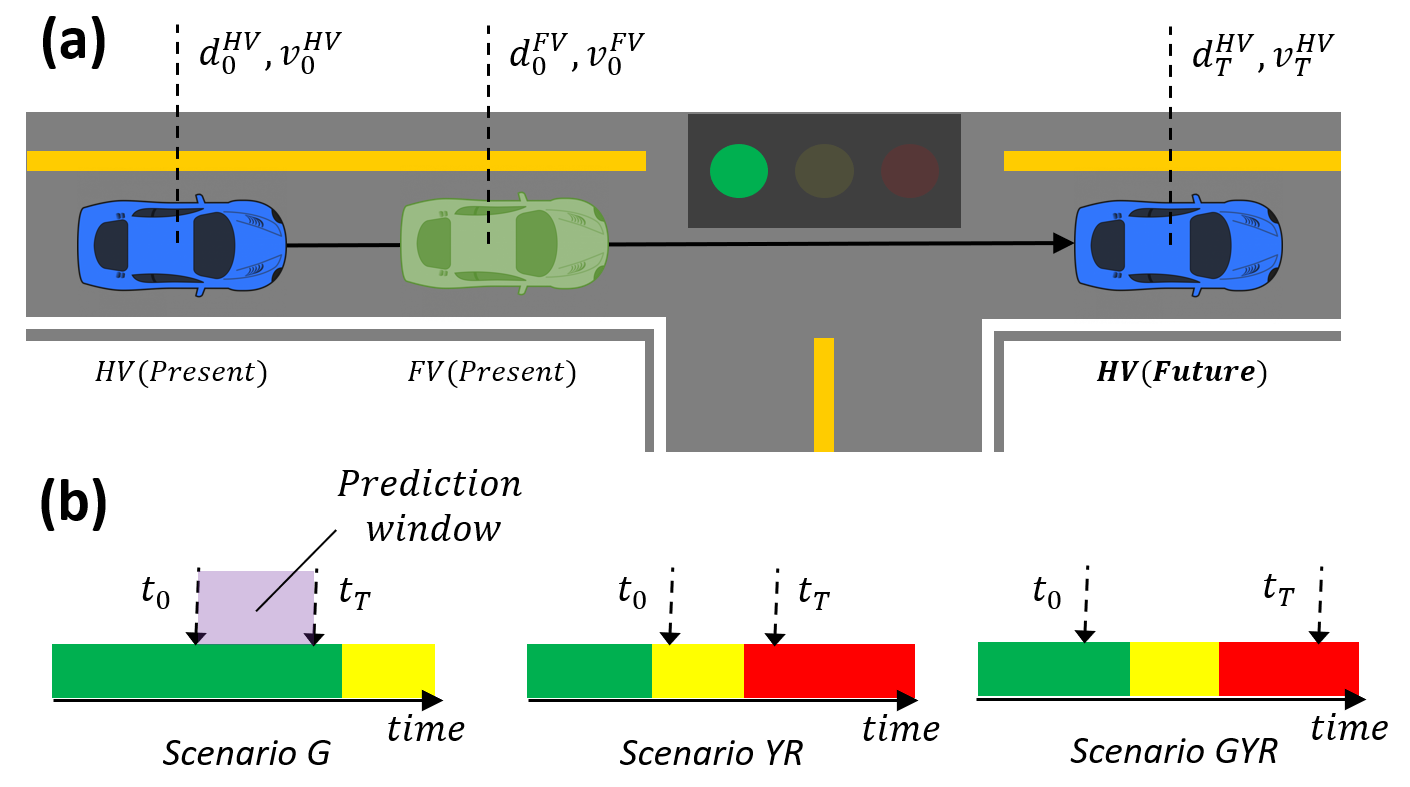}
    \vspace*{-6mm}
    \caption{(a) The trajectory forecasting problem near a traffic light for the vehicles with \textit{through} moves is depicted. Given a sequence of past states of a host vehicle (HV) and contexts (states of TLs, its front vehicle (FV)), our goal is to forecast HV's states under various states of TLs. (b) Three example scenarios of the problem are described. The full list of the scenarios is described in Table.~\ref{Table:Scenarios}. We define ‘scenario G’ as a forecasting problem when the prediction window starts on a green light and ends on the same green light. 'Scenario GYR’ represents a forecasting problem where the window spans over a set of green, yellow, and red lights.}
    \label{Fig:Intro}
\end{figure}

\begin{figure}[t] 
    \centering
    \includegraphics[width=0.78\linewidth]{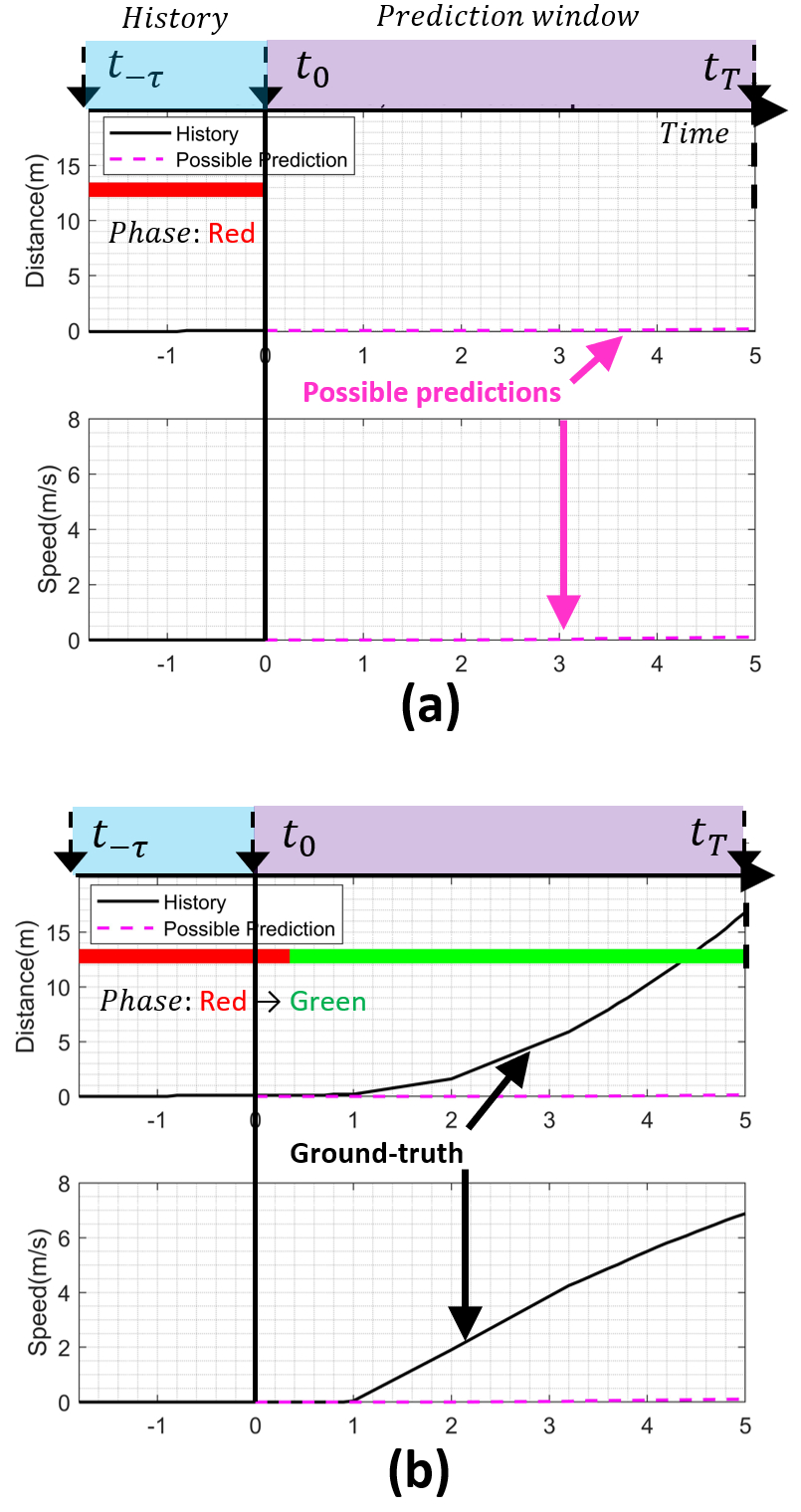}
    \vspace*{-2mm}
    \caption{A motivational example from our real-world dataset is depicted to demonstrate how uncertainty in the states of TLs makes the trajectory forecasting difficult. The sample is a 7s long RG scenario, where the first 2s is the observed history and the last 5s is the future. The goal is to predict \emph{longitudinal} position and speed of a vehicle for $t=0s:5s$. We define $0m$ as the position of the vehicle at $t=0s$. The distance to TL is then $13m$. (a) Given the history of the vehicle's position, speed, and the red phase observed at $t=0s$, a reasonable prediction under uncertain traffic phase at $t>0s$ that an existing method would make is to predict the vehicle to stay put (pink dotted line). (b) However, in reality, the phase shifted to green at $t=0.35s$ and the vehicle sped away as depicted as black line, resulting in large prediction errors in both position and speed.}
    \label{Fig:Motivational_Example}
\end{figure}

The less recognized impact of TLs in the vehicle trajectory forecasting is elaborated in Fig.~\ref{Fig:Motivational_Example}. Specifically, Fig.~\ref{Fig:Motivational_Example} depicts an example which shows how the states of TLs affect vehicle trajectories and how uncertainties in the states of TLs can cause high prediction errors. Even for models that account for the uncertainties in the future phase (it can either be red or green) and output probabilistic predictions for the two possibilities (either the phase remains red, or the phase shifts to green), the uncertainty question still is not resolved: precisely when will the phase change?

On the other hand, there have been efforts to model the dynamic impact of TLs in the transportation research community. However, there is no comprehensive model that describes behavior of human drivers near traffic signals. A few papers have studied specific instances of the dynamics but limited to a few simple scenarios; \cite{ref:1995_A0_Bennett, ref:2005_A0_Wang, ref:2018_Modified_IDM} developed models for vehicles approaching a signalized intersection (SI) and making complete stops in red light. \cite{ref:2018_Modified_IDM, ref:1987_D0_Akcelik, ref:2002_D0_Bham, ref:1990_D0_ATL, ref:2013_D0_dey} proposed models for vehicles departing from a SI from zero-speed in green phase. These models are either limited to specific instances of the problem, or are not forecasting algorithms since they require parameters like total deceleration time, final speed, which can only be measured after a trip is complete. \cite{ref:2011_highway_Park, ref:2017_highway_Jiang} presented prediction algorithms for the vehicles in highway and in car-following scenarios \cite{ref:2015_CarFollowing1_Fadhloun} based on car models proposed in \cite{ref:2000_IDM, ref:1981_Gipps}. These models, however, do not describe how human drivers react to the traffic signals.

In order to leverage rich information that comes from the dynamics between TLs and human drivers and to apply it to the trajectory forecasting problem, we simply utilize vehicle-to-infrastructure (V2I) communications. By leveraging V2I communication, one can access \textbf{the future profile} of the TLs ahead of time, which resolves the aforementioned uncertainty problem inherent to any prediction task near traffic lights. Based on this simple but novel idea, we investigate how the access to the future profile of the TLs can improve the accuracy of the trajectory forecasts.

In this paper, we tackle a trajectory forecasting problem described in Fig.~\ref{Fig:Intro}. In order to solely focus on the impact of TLs, the forecasting task is simplified to a longitudinal trajectory forecasting with a front vehicle (FV) near a TL. This setting assumes that the impact of rear \& side vehicles is minimal and no vehicle cuts into the lane that HV is located. 

Our solution approach consists of two models which map a sequence of states of a HV, its FV and the corresponding states of a TL to the subsequent action (longitudinal acceleration) that the HV takes. We name the models \emph{Human Policy Models} to highlight that the models return an action given a state. Since the models are learned to mimic drivers, the proposed approach can be viewed as \emph{imitation learning}, or precisely \emph{behavior cloning}, as we work with real data (i.e., no access to environment) and assume that inputs are i.i.d.

The first model is \emph{deterministic} human policy model which returns the most-probable action and is designed using RNN-based network. The second model is \emph{probabilistic} (generative) human policy model which outputs distribution parameters and is able to generate trajectories by sampling from the distribution and is designed using a RNN-based mixture density network (MDN) \cite{ref:1994_MDN_Bishop}. Then, we utilize these models to forecast longitudinal trajectories of the HV sequentially over a time span using the system dynamics (trasition function). For the training, validation, and testing, we used 502,253 sequences excerpted from naturalistic (non-obstructive, uncontrolled) trips from 50 distinct cars over 2 years at a signalized intersection in Ann Arbor, MI.

The remainder of the paper is organized as follows: Section II elaborates on the proposed human policy models. Section III describes the framework is used to obtain the forecast trajectories. Section IV presents results using an ablation study. Finally, Section V offers concluding remarks.

\section{Human Policy Model}
\label{section:2}
\subsection{Problem Description}
The problem of longitudinal trajectory forecasting under various phases and timings of TLs (Fig.~\ref{Fig:Intro}) is challenging due to the stochastic reactions of human drivers. For example, a driver may prefer late hard-braking approaching to a red light, while others may prefer soft-braking. Also, the reactions of drivers at steady phases (i.e., green (G), yellow (Y), red (R)) are different from those at phase transitions (GY, YR, RG). Another example is the decision making in yellow light dilemma zone \cite{ref:2007_YDilemma_Analysis_Gates}, where a driver arrives at a TL at a high speed. There usually exists two competing decisions; a driver could either engage in a hard-braking to stop before the TL or pass through the TL. 

In this sense, we break down the problem to six distinct scenarios depicted in Fig.~\ref{Fig:Intro}(b) and Table.~\ref{Table:Scenarios}. The idea behind this categorization is our belief that humans react differently to various phases of TLs, resulting in unique trajectories. In this regard, we argue that a \emph{comprehensive} model should be validated against all the 6 scenarios.

\subsection{Related Works}
To the best of our knowledge, none of the existing works in machine learning community addressed this particular forecasting problem (see Section~\ref{sec:intro}). In traditional transportation community, a few papers have discussed acceleration models or speed profiles near traffic signals. \cite{ref:2018_Modified_IDM, ref:2002_D0_Bham, ref:1990_D0_ATL, ref:2013_D0_dey} proposed polynomial speed \& acceleration models for vehicles departing from a SI from zero-speed in G phase. \cite{ref:1995_A0_Bennett, ref:2005_A0_Wang, ref:2018_Modified_IDM} developed deceleration models for vehicles make complete stops at a TL in R phase. However, these models studied very specific instances of the problem, thus do not qualify as  comprehensive model. We classified the available studies in Table.~\ref{Table:Scenarios}.

\begin{table}[h]
\caption{Six distinct scenarios of the prediction problem}
\label{Table:Scenarios}
\begin{tabular}{
>{\centering\arraybackslash}m{0.9cm} 
>{\centering\arraybackslash}m{2.0cm}
>{\centering\arraybackslash}m{4.5cm} }
\toprule
\multicolumn{2}{c}{\textbf{Scenario}} & \textbf{Available Studies  \hspace{60pt} in Transportation Community}  \\
\toprule
\multirow{2}{*}{G} & D0 (departure from zero-speed) & ATL Newzealand(1990), Bham(2002), Day(2013), Modified IDM(2018)\\
\cmidrule{2-3}
& General & None \\
\midrule
Y & \multicolumn{2}{c}{None} \\
\midrule
\multirow{2}{*}{R} & A0 (arrival to zero-speed) & Bennett(1995), Wang(2005), Modified IDM(2018) \\
\cmidrule{2-3}
& General & None \\
\midrule
GY & \multicolumn{2}{c}{None} \\
\midrule
YR & \multicolumn{2}{c}{None} \\
\midrule
RG & \multicolumn{2}{c}{None} \\
\bottomrule
\end{tabular}
\vspace*{-2pt}
\end{table}

We believe that a general model which captures the behavior of human drivers in all scenarios described in Table.~\ref{Table:Scenarios} is crucial to accurate forecasting of human vehicles near TLs. To the best of our knowledge, such model does not exist.

\subsection{Proposed Model}
The key behind modeling a driver's reaction to TL is feature selection and model design so that the reaction to TL is well captured in corresponding state space. We first introduce the model, define the state, and move on to the explanation of the intuition behind the selections. 

Human Policy Models are functions that map a sequence of past states of a HV ($X^{HV}_{t-\tau:t}:=[d_{t-\tau:t},v_{t-\tau:t}]$) and a context vector ($C_{t}:=[X^{FV}_{t}, X^{TL}_{t}, TOD_{t}]$) to a subsequent longitudinal acceleration of HV ($a^{HV}_{t}$), where $d_{t}$ and $v_{t}$ indicate the distance to the traffic light and speed at time $t$. $C_{t}$ includes state of the FV at time $t$ ($X^{FV}_{t}:=[r_{t},\dot{r}_{t}]$) defined as positions and speed relative to the HV, state of the corresponding TL at time $t$ ($X^{TL}_{t}:=[P_{t}, T_{t}]$) defined as phase (G,Y,R) as $P_{t}$ and timing (i.e., time elapsed in the current phase) of TLs as $T_{t}$, and time of day at time $t$ ($TOD_{t}$). The intuitions behind the selection of the input features ($X_{t}:=[X^{HV}_{t-\tau:t}, X^{FV}_{t}, X^{TL}_{t}, TOD_{t}]$) are as follows. \\

\noindent \textbf{Distance to traffic light} ($d_{t}$), \textbf{Speed} ($v_{t}$)
each represents the longitudinal distance of a HV to the TL that the HV is approaching or departing from and the longitudinal speed of the HV. They are essential in forecasting vehicle trajectories near TLs. For example, a HV approaching a TL in red phase travels slowly when it is close to the TL, whereas it can travel fast when it is far away from the TL. $d_{x} > 0$ means that the HV is approaching the TL, and $d_{x} < 0$ indicates that it's departing from the TL. We assume $v_{x} >= 0$.

\noindent \textbf{Range and range-rate} ($r_{t}, \dot{r}_{t}$) represent the longitudinal position and speed of the FV relative to those of the HV. We assume $r_{x} > 0$, meaning that the FV is always ahead of the HV. Note, rear \& side vehicles are not considered as we simplified the problem to concentrate on the impact of TL. However, one can trivially extend our model to include them. 

\noindent \textbf{Phase and timing of traffic light} ($P_{t}, T_{t}$)
represents the phase of a TL (G,Y,R) that a HV is subject to and the time elapsed since the last phase change ($T_{t}>=0$). $T_{t}$ accounts for transient behaviors of human drivers at phase shifts. For example, a vehicle approaching a TL in a red phase with a small $T_{t}$, meaning that the phase just shifted to red, may not be traveling slowly whereas a vehicle approaching a TL with a large $T_{t}$ is likely to travel slowly or fully stopped. 

\noindent \textbf{Time of day ($TOD$)}
represents the time of day as elapsed hours since midnight ($0\leq TOD<24$). TOD=12 indicates noon. Macroscopic traffic characteristics including traffic speed differ considerably depending on $TOD$ as evidenced in studies including \cite{ref:VED_Oh}. The selection of $TOD$ is an attempt to incorporate the macroscopic trend of the traffic. \\

Due to stochastic and complex nature of human decision making in driving, a simple analytical model such as a polynomial or a physics-based model may not represent the nominal or probabilistic behaviors of human-drivers near traffic signals well. This is why we opt for data-driven approach.

\subsection{Dataset}
\label{sec:dataset}
The real-world driving data utilized in this work are from Safety Pilot Model Deployment (SPMD), a large-scale connected vehicle study conducted in Ann Arbor, MI \cite{ref:SPMD}. The vehicles were equipped with data loggers which collected 10Hz GPS signals including coordinates, speeds, and heading angles as well as 10Hz front vehicles data such as relative positions and speeds. While SPMD database does not include any vision data (e.g., lidar, camera, or radar), it provides detailed TL profiles that were obtained using V2I communication devices installed at signalized intersections (SIs). To the best of our knowledge, no dataset is publicly available that provides the detailed TL data. Hence, we leverage SPMD for its unique access to the TL profiles.

In this work, we extracted 502,253 observations (samples) from 50 distinct SPMD vehicles collected over a span of 27 months (Mar 2015 - July 2017) near the Plymouth Rd \& Huron Pkwy intersection. Each observation was synchronized with the traffic signal states of the SI. In order to reduce the noise in $X^{HV}_{t}, X^{FV}_{t}, a^{HV}_{t}$, a least-square polynomial smoothing filter was used \cite{ref:SG_Filter}.


\begin{figure}[t] 
    \centering
    \includegraphics[width=0.75\linewidth]{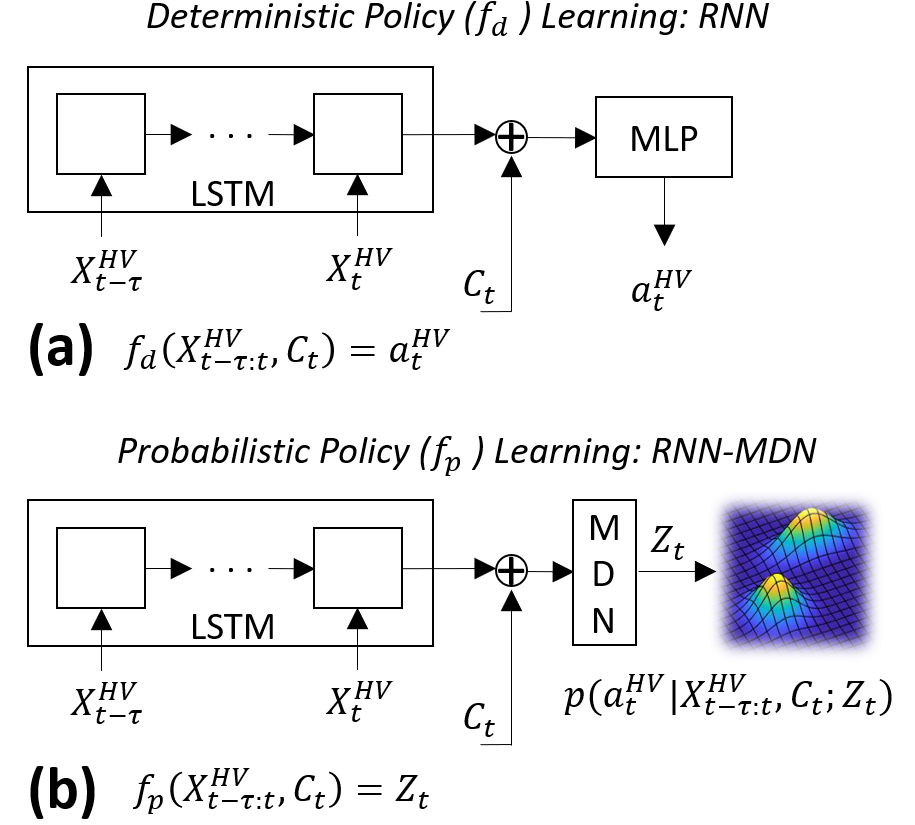}
    \vspace*{-2mm}
    \caption{The proposed policy models are described. (a) a deterministic policy ($f_{d}:[X^{HV}_{t-\tau:t}, C_{t}]\rightarrow a^{HV}_{t}$), (b) a probabilistic policy ($f_{p}:[X^{HV}_{t-\tau:t},C_{t}]\rightarrow Z_{t}$).}
    \label{Fig:methods}
\end{figure}

\subsection{Implementation Details}
Both the deterministic ($f_{d}$) and probabilistic ($f_{p}$) human policy models were implemented in Tensorflow-Keras. Each model consists of a double-stacked LSTM which takes $X^{HV}_{t-\tau:t}$ followed by a concatenation with the context vector $C_{t}=[X^{FV}_{t}, X^{TL}_{t}, TOD_{t}]$. The concatenated tensor is then fed into a multi-layer perceptron (MLP) for the deterministic model or a MDN for the probabilistic model. The MLP layer outputs $a^{HV}_{t}$ whereas the MDN layer outputs the distribution parameters $Z_{t}$. As we model MDN using gaussian mixtures, the MDN layer outputs three sets of the parameters: mixture weights $\pi_{k}$, means $\mu_{k}$, variances $\sigma^{2}_{k}$ for $N$ components. We used $N=2$ for the yellow light dilemma scenario. Models were trained using ADAM optimizer.

$f_{d}$ is learned by minimizing a loss function $L_{d}$ which is a summation of mean squared error as described below.
\begin{equation}
L_{d} := \sum_{t=1}^{T} (a^{HV}_{t}-f_{d}(X^{HV}_{t-\tau:t},C_{t}))^2
\label{Loss_d}
\end{equation}

$f_{p}$ is obtained by minimizing a loss function $L_{p}$ which is a sum of a negative log-likelihood.
\begin{equation}
L_{p} := \sum_{t=1}^{T} -log(p(a^{HV}_{t}|X^{HV}_{t-\tau:t},C_{t};Z_{t}))
\label{Loss_p}
\end{equation}

\section{Trajectory Forecast Framework}
\label{sec:method}
\begin{figure}[t] 
    \centering
    \includegraphics[width=0.9\linewidth]{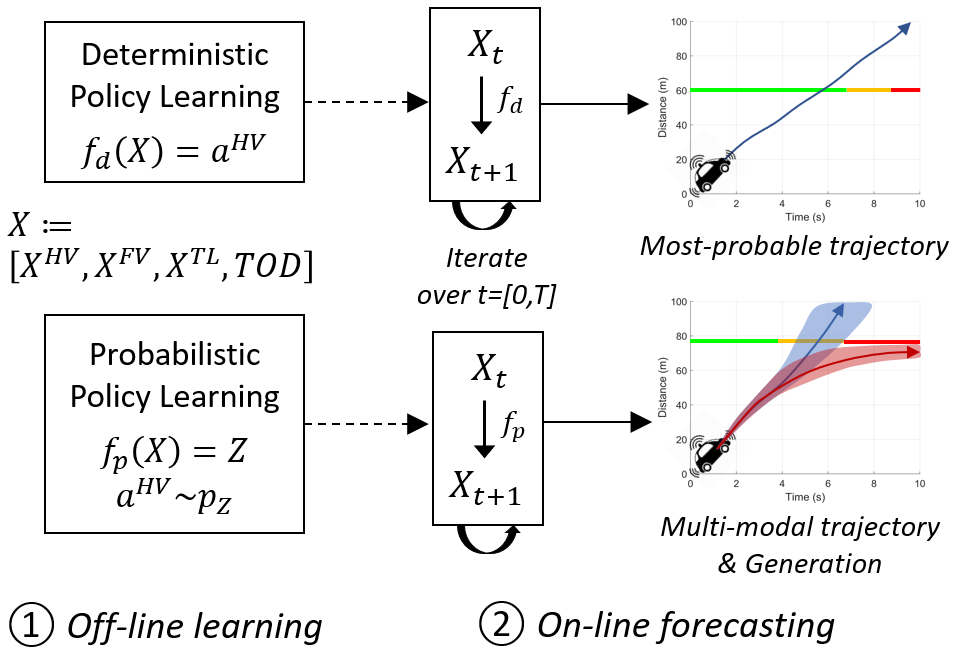}
    \vspace*{-2mm}
    \caption{The framework is divided into two steps. The first step is the off-line training of the proposed models. The second step involves propagations of the policies and state transitions to obtain vehicle trajectories over $T$.}
    \label{Fig:Framework}
\end{figure}

Fig.~\ref{Fig:Framework} illustrates the proposed framework that consists of two parts. The first part is an off-line supervised learning of the policies. The second part is where we use the learned policies in sequential prediction setting to obtain $X^{HV}_{0:T}$ (i.e., alternation of one-step predictions of the learned policies and state transitions). The system dynamics (i.e., transition function for $X^{HV}$) is given as longitudinal vehicle kinematics with a zero-order hold, as described in Eq.~\ref{Eq:next_dxvx} and \ref{Eq:Onestep_vehicle_dynamics}.
\vspace{-4mm}

\begin{equation}
v_{n+1} := v_{n} + a_{n}\varDelta{t}_{n},   d_{n+1} := d_{n} + 0.5(v_{n+1}+v_{n})\varDelta{t}_{n}
\label{Eq:next_dxvx}
\vspace*{-2mm}
\end{equation}

\begin{equation}
\label{Eq:Onestep_vehicle_dynamics}
X^{HV}_{n+1}=A_{n}X^{HV}_{n}+B_{n}a^{HV}_{n}
\end{equation}

where $A_{n} := \begin{bmatrix} 1 & \varDelta{t}_{n}\\ 0 & 1 \end{bmatrix}, B_{n} := \begin{bmatrix} 0.5\varDelta{t}_{n} \\ \varDelta{t}_{n} \end{bmatrix},  X^{HV}_{n} = \begin{bmatrix} d_{n} \\ v_{n} \end{bmatrix}. \\ $

Defining $n=0$ and $n=N$ as the index for $t=0$ and $t=T$, the trajectory forecast over the prediction horizon $t=[0,T]$ are obtained by propagating Eq.~\ref{Eq:Onestep_vehicle_dynamics} from $n=0$ to $n=N-1$:

\vspace{-6pt}
\begin{equation}
\label{Eq:Multistep_vehicle_dynamics}
\begin{aligned}
X^{HV}_{N}
= \prod^{N-1}_{k=0}A_{k}X^{HV}_{0}
+ \prod^{N-1}_{k=1}A_{k}B_{0}a^{HV}_{0}
+ \prod^{N-1}_{k=2}A_{k}B_{1}a^{HV}_{1}
\\[1pt]
+ ... 
+ A_{N-1}B_{N-2}a^{HV}_{N-2}
+ B_{N-1}a^{HV}_{N-1}
\\[1pt]
= \prod^{N-1}_{k=0}A_{k}X^{HV}_{0}
+ \prod^{N-1}_{k=1}A_{k}B_{0}f(X^{HV}_{-n_{\tau}+1:0}, C_{0})
\\[1pt]
+ ... 
+ B_{N-1}f(X^{HV}_{N-n_{\tau}:N-1}, C_{N-1})
\\[1pt]
:= F(X^{HV}_{1:N-1},X^{HV}_{-n_{\tau}+1:0},C_{0:N-1},\varDelta{t}_{0:N-1})
\vspace*{-6mm}
\end{aligned}
\end{equation}

where $f$ can either be $f_{d}$ or $S(f_{p})$. $S$ is a function which returns a sample $a^{HV}_{t}$ from the pdf. In case of 1D Gaussian, $Z_{t}:=[\mu_{t}, \sigma^{2}_{t}]$ and $a^{HV}_{t} \sim N(\mu_{t}, \sigma^{2}_{t})$. $\tau, n_{\tau}$ each indicates input sequence length in time and in the number of steps. 

As described in Eq.~\ref{Eq:Multistep_vehicle_dynamics}, $X^{HV}_{N}$ is a function ($F$) of [$X^{HV}_{1:N-1}$, $X^{HV}_{-n_{\tau}+1:0},C_{0:N-1},\varDelta{t}_{0:N-1}$]. The second term $X^{HV}_{-n_{\tau}+1:0}$ is given and the last term $\varDelta{t}_{0:N-1}$ can simply be predetermined based on a required time resolution. Obtaining $C_{0:N-1}$ at the prediction time ($t=0$) is the main challenge, due to uncertainties in $X^{FV}_{1:N-1}, X^{TL}_{1:N-1}$. A simple way to get away with the uncertainties is to design a model to predict trajectories ($X^{HV}_{0:T}$) conditioned only on the observed states [$X^{FV}_{-\tau:0}, X^{TL}_{-\tau:0}$]. An example is a model with many-to-many RNN that takes a sequence of past states and returns a sequence of future states; which is a forecasting model that does not utilize the future states of TLs. Fig.~\ref{Fig:Motivational_Example} showed how such model can fail.

\begin{figure*}[t] 
    \centering
    \includegraphics[width=0.96\linewidth]{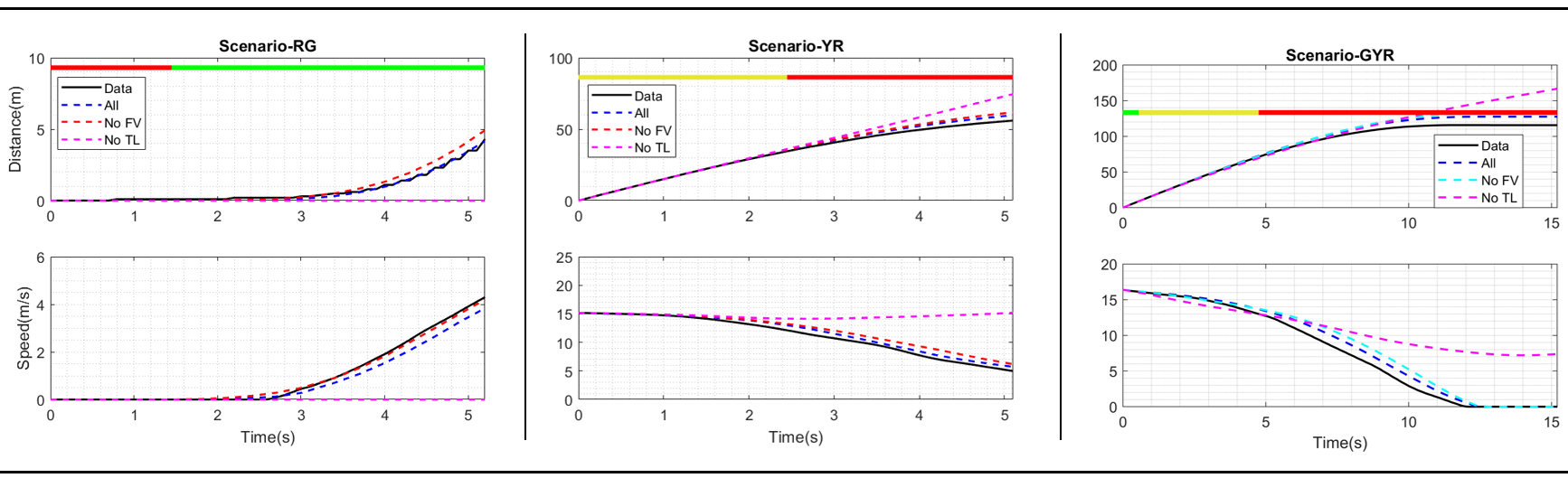}
    \vspace{-2mm}
    \caption{For qualitative evaluations, ground-truth and predicted trajectories are depicted for two 5s sample scenarios RG (left), YR (middle), and one 15s sample scenario GYR (right). Four trajectories of distance to the TL (1st row), speed (2nd row) are shown in each plot and represent ground-truth (black), predictions $X^{HV}_{0:5s}$ obtained using $f_{d}$(Blue), $f^{NoFV}_{d}$(Red), $f^{NoTL}_{d}$(Pink). The 2s history $X^{HV}_{-2:0s}$ (i.e., part of inputs to the models) are omitted for the simplicity.}
    \label{Fig:Visual_DASFV}
\end{figure*}

This is where our novel idea comes into play. We remove the uncertainties by utilizing the future phases and timings of TLs obtained via V2I communications. With the access, $X^{TL}_{1:N-1}$ can be attained at the prediction time. The remainder is then $X^{FV}_{1:N-1}$, which is obtained using a variant of $f_{d}$. Specifically, we train another human policy model $f^{NoFV}_{d}:[X^{HV}_{N-n_{\tau}:N}, C^{\prime}_{N}] \rightarrow a^{HV}_{N}$ with $C^{\prime}_{N}:=[X^{TL}_{N}, TOD_{N}]$ excluding $X^{FV}_{N}$ from $C$ (i.e., $f^{NoFV}_{d}$ does not consider FV). After the off-line learning of $f^{NoFV}_{d}$, we apply the aforementioned iterative process on $FV$ to obtain $X^{FV}_{1:N-1}$ via Eq.~\ref{Eq:Multistep_vehicle_dynamics} with $f^{NoFV}_{d}$. 

Once $X^{FV}_{1:N-1}, X^{TL}_{1:N-1}$ are secured, the resulting trajectory forecast $X^{HV}_{1:N}$ is obtained via Eq.~\ref{Eq:Multistep_vehicle_dynamics}. Since $X^{HV}_{1:N}$ can simply be forecast using $f^{NoFV}_{d}$, we conduct an ablation study (elaborated in Section~\ref{section:results}) on $f_{d}$, $f^{NoFV}_{d}$ and the other two models ($f^{NoTL}_{d}, f^{NoFVTL}_{d}$) which each represents unique model where $X^{TL}$ and $[X^{FV}, X^{TL}]$ are excluded from $C$.

For the probabilistic human policy model, the probability of the resulting trajectory forecast $p(X^{HV}_{1:N})$ can be estimated using the chain rule of probability, which factorizes the joint distribution over $N$ separate conditional probabilities:

\begin{equation}
\label{Eq:prob_estimation}
\begin{aligned}
\\[-12pt]
p(X^{HV}_{1:N}|X^{HV}_{-n_{\tau}+1:0},C_{0:N-1},\varDelta{t}_{0:N-1}) =
\\[1pt]
\prod^{N}_{k=1}p(X^{HV}_{k}|X^{HV}_{1:k-1},X^{HV}_{-n_{\tau}+1:0},C_{0:k-1},\varDelta{t}_{0:k-1})
\end{aligned}
\end{equation}

As opposed to the deterministic forecasting where the most-probable trajectory is obtained, a resulting trajectory is a sample from a probability distribution. While we can estimate the joint probability density of a trajectory forecast via Eq.~\ref{Eq:prob_estimation}, the marginal probability of $X^{HV}_{t}$ needs to be numerically estimated via sampling since the distribution parameter $Z_{t}$ is obtained via an arbitrarily complex neural network and depends on previous predictions $X^{HV}_{-n_{\tau}+1:k-1}$. Thus, we utilize Monte Carlo Simulation to obtain the samples (roll-out trajectories) and kernel density estimation to approximate the marginal probability density of the samples.

\section{Results}
\label{section:results}

In this section, we discuss evaluation results conducted on the dataset (see Section~\ref{sec:dataset}). In Section~\ref{sub:Results1}, we present the impact of $X^{TL}$ qualitatively by comparing the trajectory forecasts made using the four deterministic policies and that the utilization of phases and timings of TLs helps forecasting trajectories more accurately. In Section~\ref{sub:Results2}, we discuss a set of metrics to evaluate performance of the forecasts. The ablation study in Section~\ref{sub:Results3} is designed to evaluate the 4 variants of our deterministic models ($f_{d}, f^{NoFV}_{d}, f^{NoTL}_{d}, f^{NoFVTL}_{d}$) by quantifying their performance using 3,111 test snippets. In Section~\ref{sub:Results4}, we demonstrate how the proposed probabilistic models can tackle scenarios with competing policies.

\subsection{Impact of $X^{TL}$ on trajectory forecasts}
\label{sub:Results1}
As explained in Section~\ref{sec:method}, we first train 4 distinct deterministic policy models $f_{d}, f^{NoFV}_{d}, f^{NoTL}_{d}$, and $f^{NoFVTL}_{d}$. The superscript specifies the context input $C$ that the model takes (i.e., $C^{NoFVTL}:=[TOD], C^{NoTL}:=[X^{FV},TOD], C^{NoFV}:=[X^{TL},TOD]$, where $C^{mode}$ is the context input for $f^{mode}_{d}$). Each model then produces unique trajectory forecasts per scenario. Three scenarios (RG, YR, GYR) are sampled and the four models are used to forecast trajectories. In each sample, four trajectories are depicted to represent ground-truth and 3 predicted trajectories each from $f_{d}, f^{NoFV}_{d}$, $f^{NoTL}_{d}$.



\begin{figure*}[t] 
    \centering
    \includegraphics[width=0.91\linewidth]{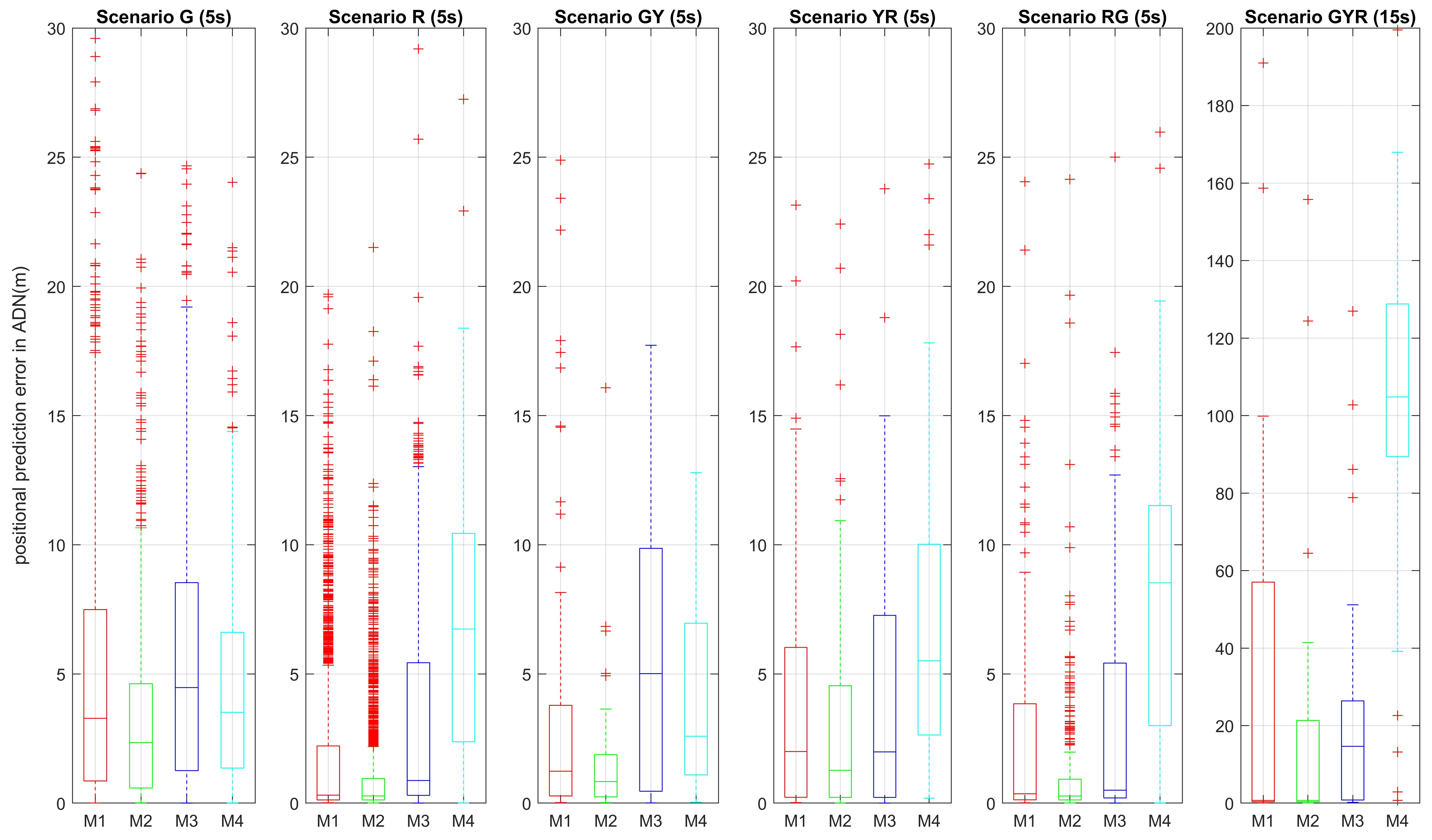}
    \vspace{-3mm}
    \caption{A quantitative evaluation is conducted on the test set and depicted as boxplots for the six scenarios using the evaluation metric \textit{ADN}. M1, M2, M3, M4 each represents forecasting models with $f_{d}, f^{NoFV}_{d}, f^{NoTL}_{d}, f^{NoFVTL}_{d}$. The prediction window is $T=15s$ for scenario GYR, and $T=5s$ for others.}
    \label{Fig:Boxplot}
\end{figure*}

The sample on the left of Fig.~\ref{Fig:Visual_DASFV} is similar to the motivational example in Fig.~\ref{Fig:Motivational_Example}. At $t=0$, the driver was at a stop at a red phase. A reasonable prediction that a model without $X^{TL}_{0:5s}$ would make is to forecast the vehicle to stay put. As expected, $f^{NoTL}_{d}$ (pink) failed to predict $X^{HV}$ accurately, whereas the forecasts from the other two models $f_{d}, f^{NoFV}_{d}$ which utilize $X^{TL}_{0:5s}$ are close to the ground-truth. 

The middle plot of Fig.~\ref{Fig:Visual_DASFV} is a scenario YR where the driver was cruising as approaching the TL ($v^{HV}_{-2:0s}=15$). Given $P_{0}=Y$ and that the vehicle was cruising, $f^{NoTL}_{d}$ forecast the vehicle to maintain the speed, causing the prediction errors to grow over time. The other two models $f_{d}, f^{NoFV}_{d}$ which use $X^{TL}_{0:5s}$ took account for the phase shift in the future and accurately forecast how the driver would react to the shift.

On the other hand, the right plot serves as an exemplar long-term scenario. A 15s scenario that spans a full cycle of phases (GYR) is illustrated where HV was initially decelerating while approaching the TL. All models captured temporal trends in the speed and predicted that HV would continue to decelerate. $f^{NoTL}_{d}$ predicted that the vehicle would cross the intersection, given $P_{0}=G$ whereas $f_{d}, f^{NoFV}_{d}$ predicted that the vehicle would make a stop before the TL considering the future states of TL. Indeed, the ground-truth is that the vehicle made a stop before the TL to react to the phase shift.

As demonstrated, the impact of $X^{TL}$ is significant; uncertainties in $X^{TL}$ can cause high prediction errors, especially for long-term predictions. The results suggest that forecasting methods may perform poor without knowledge of future $X^{TL}$, highlighting why the problem is critical. While information of TLs can be transmitted through FV to some extent, such transmission is ineffective when (1) FV is absent, (2) FV is present, but, far from HV, or (3) FV is present and close to HV, however, the phase transition is imminent. The proposed idea is a solution to the problem: utilization of the future $X^{TL}$ greatly improves the quality of forecasts near traffic lights.


\subsection{Evaluation metrics}
\label{sub:Results2}
We use the following metrics for the quantitative evaluation: mean absolute error (MAE), time weighted absolute error (TWAE), absolute deviation at the end of the prediction window (ADN) defined in Eq.~\ref{Eq:MAE}, \ref{Eq:TWAE}, and \ref{Eq:ADN}, where $\hat{X}^{HV}_{k}, X^{HV}_{k}$ represents the $k$th-step forecast $X$ and ground truth $X$.

\begin{equation}
MAE := \frac{\sum_{k=1}^{N} |\hat{X}^{HV}_{k}-X^{HV}_{k}|}{N}
\label{Eq:MAE}
\vspace*{-1mm}
\end{equation}

\begin{equation}
TWAE := \frac{\sum_{k=1}^{N} (t_{k}|\hat{X}^{HV}_{k}-X^{HV}_{k}|)}{\sum_{k=1}^{N}t_{k}}
\label{Eq:TWAE}
\vspace*{-1mm}
\end{equation}

\begin{equation}
ADN := |\hat{X}^{HV}_{N}-X^{HV}_{N}|
\label{Eq:ADN}
\end{equation}

We used $\forall k: \varDelta{t_{k}}=0.2s$, $\tau=2s$ (history). For the scenario with a prediction window $t_{N}=5s$, the last index $N$ is 25.


\begin{table*}[h]
\caption{Ablation study on \textbf{position} errors with average MAE, TWAE, ADN. Prediction horizon is 15s for GYR and 5s for others. The lower a metric is the better. The numbers from the best performing model are marked in \textbf{bold}. }
\label{Table:ablation_study_table}
\begin{tabular}{M{1.5cm}|M{0.9cm}M{0.9cm}M{0.9cm}M{0.9cm}|M{0.9cm}M{0.9cm}M{0.9cm}M{0.9cm}|M{0.9cm}M{0.9cm}M{0.9cm}M{0.9cm}}
\hline \T \B

\multirow{2}{*}{\textbf{Scenario}} & 
\multicolumn{4}{c}{\textbf{MAE(m)}} & 
\multicolumn{4}{c}{\textbf{TWAE(m)}} & 
\multicolumn{4}{c}{\textbf{ADN(m)}} \\
\cline{2-13} \T \B

& 
All & NoFV & NoTL & NoFVTL &
All & NoFV & NoTL & NoFVTL &
All & NoFV & NoTL & NoFVTL \\
\cline{1-13}

& 
 &  &  &  &
 &  &  &  &
 &  &  &  \\[-3pt]

G & 
1.18 & \textbf{0.78} & 1.35 & 1.19 &
1.86 & \textbf{1.21} & 2.16 & 1.78 &
3.28 & \textbf{2.34} & 4.47 & 3.51 \\[2.5pt]

R & 
0.16 & \textbf{0.15} & 0.38 & 1.79 &
0.21 & \textbf{0.20} & 0.56 & 2.98 &
0.31 & \textbf{0.28} & 0.88 & 6.74 \\[2.5pt]

GY & 
0.53 & \textbf{0.41} & 1.26 & 0.82 &
0.77 & \textbf{0.54} & 2.06 & 1.31 &
1.24 & \textbf{0.84} & 5.01 & 2.59 \\[2.5pt]

YR &
1.18 & \textbf{0.69} & 0.86 & 1.01 &
1.60 & \textbf{0.93} & 1.32 & 1.42 &
2.00 & \textbf{1.27} & 1.98 & 2.29 \\[2.5pt]

RG &
0.17 & \textbf{0.14} & 0.22 & 2.28 &
0.24 & \textbf{0.19} & 0.31 & 3.80 &
0.36 & \textbf{0.28} & 0.50 & 8.53 \\[2.5pt]

GYR &
\textbf{0.445} & \textbf{0.445} & 4.11 & 32.40 &
\textbf{0.567} & 0.568 & 6.82 & 51.16 &
\textbf{0.694} & \textbf{0.694} & 14.65 & 104.81 \\[2.5pt]

\hline
\end{tabular}
\end{table*}

\begin{table*}[h]
\caption{Ablation study on \textbf{velocity} errors with average MAE, TWAE, ADN. Prediction horizon is 15s for GYR and 5s for others. The lower the better.}
\label{Table:ablation_study_table_velocity}
\begin{tabular}{M{1.5cm}|M{0.9cm}M{0.9cm}M{0.9cm}M{0.9cm}|M{0.9cm}M{0.9cm}M{0.9cm}M{0.9cm}|M{0.9cm}M{0.9cm}M{0.9cm}M{0.9cm}}
\hline \T \B

\multirow{2}{*}{\textbf{Scenario}} & 
\multicolumn{4}{c}{\textbf{MAE(m/s)}} & 
\multicolumn{4}{c}{\textbf{TWAE(m/s)}} & 
\multicolumn{4}{c}{\textbf{ADN(m/s)}} \\
\cline{2-13} \T \B

& 
All & NoFV & NoTL & NoFVTL &
All & NoFV & NoTL & NoFVTL &
All & NoFV & NoTL & NoFVTL \\
\cline{1-13}

& 
 &  &  &  &
 &  &  &  &
 &  &  &  \\[-3pt]

G & 
0.70 & \textbf{0.46} & 0.97 & 0.75 &
1.02 & \textbf{0.65} & 1.41 & 1.07 &
1.48 & \textbf{0.94} & 2.28 & 1.51 \\[2.5pt]

R & 
\textbf{0.005} & \textbf{0.005} & 0.17 & 1.44 &
0.007 & \textbf{0.006} & 0.22 & 2.29 &
0.010 & \textbf{0.009} & 0.29 & 4.10 \\[2.5pt]

GY & 
0.39 & \textbf{0.15} & 1.11 & 0.53 &
0.54 & \textbf{0.20} & 1.78 & 0.79 &
0.87 & \textbf{0.27} & 3.08 & 1.21 \\[2.5pt]

YR &
\textbf{0.367} & 0.374 & 0.53 & 1.01 &
\textbf{0.47} & 0.52 & 0.71 & 1.42 &
\textbf{0.59} & 0.69 & 0.89 & 2.29 \\[2.5pt]

RG &
0.007 & \textbf{0.005} & 0.10 & 1.68 &
0.010 & \textbf{0.007} & 0.14 & 2.63 &
0.02 & \textbf{0.01} & 0.22 & 4.52 \\[2.5pt]

GYR &
0.015 & \textbf{0.014} & 1.26 & 6.78 &
\textbf{0.018} & 0.020 & 1.86 & 9.80 &
0.030 & \textbf{0.029} & 2.60 & 14.13 \\[2.5pt]

\hline
\end{tabular}
\end{table*}


\subsection{Ablation Study}
\label{sub:Results3}
The goal of the ablation study is to quantify the impacts of $X^{TL}$ on a large testset and to evaluate performance of the four policies. The result (on ADN) is presented in Fig.~\ref{Fig:Boxplot}. The sample sizes are 688 (G), 1909 (R), 68 (GY), 81 (YR), 362 (RG), 32 (GYR), totalling 3,111 sample snippets. The detailed results for the ablation study on all three metrics (MAE, TWAE, ADN) are presented in Table~\ref{Table:ablation_study_table}, \ref{Table:ablation_study_table_velocity}, each for position and velocity errors. Note, scenario Y is not depicted due to its short (and inconsistent) prediction horizon; we observed that the phase Y usually lasts anywhere between 2.5s to 4s. The first 2s are used as inputs, which means the prediction horizon for the scenario Y is only 0.5s to 2s. 

As depicted in Table~\ref{Table:ablation_study_table}, \ref{Table:ablation_study_table_velocity}, the magnitude of error is as follows: MAE$<$TWAE$<$ADN, making ADN the largest error. As shown in Fig.~\ref{Fig:Boxplot}, across all scenarios, the two models $f_{d}, f^{NoFV}_{d}$ which utilizes $X^{TL}$ outperform the other two models $f^{NoTL}_{d}, f^{NoFVTL}_{d}$ which don't take advantage of the future phases and timings information. Interestingly, the winner is not $f_{d}$, but it is $f^{NoFV}_{d}$, which performs the best on all characteristics of boxplot including the 1st, 3rd quartiles, the median, and the upper limit of the extreme points. Our interpretation is that the exclusion of $X^{FV}$ from $C$ increases the prediction accuracy, due to the uncertainty in $\hat{X}^{FV}_{t>0}$ as $\hat{X}^{FV}_{t>0}$ are predicted via another human policy model.

The numbers presented in Table~\ref{Table:ablation_study_table}, \ref{Table:ablation_study_table_velocity} agree with the results from Fig.~\ref{Fig:Boxplot} across all scenarios. $f^{NoFV}_{d}$ is the winner for almost all metrics, or at least on par with $f_{d}$. In summary, the knowledge of future states of traffic lights significantly increase the accuracy of trajectory forecasts, as evidenced in the ablation studies: trajectory forecasts with the winner model have roughly 1.5-30 times smaller (position) MAE, TWAE, ADN for $T=5s$ scenarios (G, R, GY, YR, RG), and roughly 9-150 times smaller MAE, TWAE, ADN for $T=15s$ scenario (GYR), compared to trajectory forecasts via $f^{NoTL}_{d}, f^{NoFVTL}_{d}$. This discrepancy in the accuracy becomes bigger as the prediction horizon grows, especially for the long-term forecasts such as scenario GYR.

\begin{figure*}[h] 
    \centering
    \includegraphics[width=0.75\linewidth]{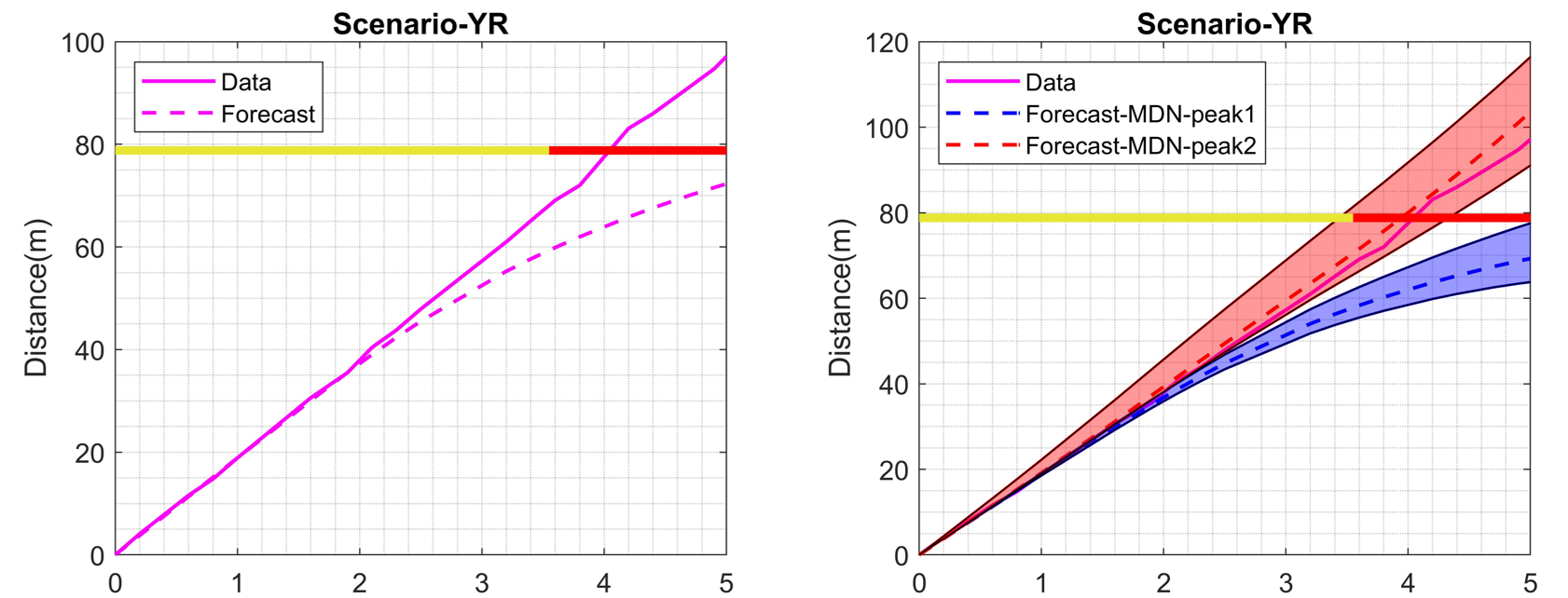}
    \vspace{-2mm}
    \caption{A sample trip for the yellow light dilemma scenario. The left plot highlights the limitation of the most-probable trajectory. The right plot shows that the proposed probabilistic models are able to forecast the two competing policies. The trajectories $\forall t: p(d_{t})>= 0.01$ are illustrated.}
    \label{Fig:Competing_policies}
\end{figure*}

\subsection{Probabilistic Prediction}
\label{sub:Results4}
\vspace{-2pt}
The outliers observed (indicated as '+') in Fig.~\ref{Fig:Boxplot} occur mostly from edge cases and competing policies. Examples of the edge cases include a driver approaching a TL in $P_{\forall t}=R$ with high speed and executing a sudden break right before the TL rather than gradually slowing down as it approaches the intersection. Another example is that a driver in the middle of the road in $P_{\forall t}=G$ moving much slower than the average speed of the traffic for unknown reasons. The outliers occur from competing policies are exemplified by the yellow light dilemma scenario where a driver can either cross the intersection or stop before the intersection.

Fig.~\ref{Fig:Competing_policies}(a) describes a sample trip observed in our dataset that represents the yellow light dilemma scenario. As shown in the figure, the most-probable trajectory forecast obtained via $f^{NoFV}_{d}$ predicts that the vehicle would make a stop before the intersection, however, the driver crossed the intersection even after the phase shifted to red.

This is where the proposed probabilistic models come in handy. As the probabilistic model is capable of reproducing multi-modal distributions, it captures the other competing policy (\emph{cross}). In addition, it is able to reason uncertainties of the forecasts as the probabilities of the forecasts can be estimated via Monte-Carlo simulation. Another advantage is that we can generate (sample) possible trajectories as the proposed probabilistic policy outputs both likelihood $p(X|Z)$ and prior $p(Z)$; i.e., the model is generative.

We argue that the deterministic models are still valuable: the solutions are simple, cost-efficient, and easy to interpret. They can serve as nominal trajectories of human drivers in situations that can be approximated uni-modal. The nominal trajectories can be used in a trajectory planning algorithm which works with deterministic actors or disturbances.

For the scenarios with 5s prediction horizon, the time to compute a most-probable trajectory forecast is less than 10milliseconds on a single-core personal laptop with i7-6500U 2.50GHz CPU, and 8GB RAM without utilizing a parallelization. However, it takes several seconds (5-10s for 1,000 rollout trajectories) to construct the pdf for the probabilistic forecasts on the same machine due. One can significantly reduce the time via parallel computing (GPU).

\section{Conclusion}
Our work is the first attempt in the community to comprehensively understand and identify the impact of traffic signals on trajectory forecasting near TLs. In this regard, we first introduced the motivational example in Fig.~\ref{Fig:Motivational_Example} and defined 6 distinct scenarios of the problem. We proposed a novel idea to solve the scenarios that leverages the \textbf{future states} of traffic lights obtained via V2I communications. Specifically, we proposed deterministic and probabilistic human policy models to simulate state-dependent driver actions near TLs. In the ablation study, we show that the utilization of future phases and timings of TLs significantly improves the quality of trajectory forecasts for all scenarios described in Table.~\ref{Table:Scenarios}.

As no dataset is publicly available that has both the detailed TL data and the vision data (camera, lidar, or radar) at the moment, our experiments were based on a non-vision dataset. Hence, a direct comparison against state-of-the-art forecasting models \cite{ref:2017_Desire, ref:2018_SGAN, ref:2019_Intention, ref:2016_SocialLSTM, ref:2019_Traphic} is not made since the models were built on the vision data. Regardless, the proposed idea (i.e., the utilization of the future states of TLs) and the proposed framework can be leveraged by any work that concerns vehicle trajectory forecasting near TLs given the access to V2I communications. As the results suggest that the utilization of the future states could lead to the significant improvements in the prediction accuracy, especially for the long-term ones, we believe that it is worth building a large-scale dataset with both the TL and vision data to quantify the influence to the fullest extent.

\section*{ACKNOWLEDGMENT}
The authors would like to thank the University of Michigan Transportation Research Institute for access to the data.


\bibliographystyle{IEEEtran}
\bibliography{LM_SI_Bib}

\end{document}